\documentclass[twoside,11pt]{article}
\usepackage{jair, theapa, rawfonts}

\jairheading{1}{2024}{1-37}{11/1}{}
\ShortHeadings{AutoSAT: Automatically Optimize SAT Solvers via Large Language Models}
{Sun et al.}

\usepackage[utf8]{inputenc} 
\usepackage[T1]{fontenc}    
\usepackage{url}            
\usepackage{booktabs}       
\usepackage{amsfonts}       
\usepackage{nicefrac}       
\usepackage{microtype}      
\usepackage{xcolor}         
\usepackage{amsmath}
\usepackage{amssymb}
\usepackage{mathtools}
\usepackage{amsthm}
\usepackage[algo2e,ruled,vlined,linesnumbered]{algorithm2e}
\usepackage{wrapfig}
\usepackage[most]{tcolorbox}
\usepackage{listings} 

\newtcolorbox{promptbox}[1][]{
  breakable,
  title=#1,
  colback=gray!5,
  colframe=black,
  colbacktitle=gray!15,
  coltitle=black,
  fonttitle=\bfseries,
  bottomrule=1.5pt,
  toprule=1.5pt,
  leftrule=1pt,
  rightrule=1pt,
  arc=0pt,
  outer arc=0pt,
  enhanced,
  before upper={\parindent=1.5em} 
}

\title{AutoSAT: Automatically Optimize SAT Solvers via Large Language Models}

\author{\name Yiwen Sun \email ywsun22@m.fudan.edu.cn
\\\addr  School of Data Science \\
  Fudan University \\
  Shanghai, China 
    \AND
  \name Furong Ye \email f.ye@ios.ac.cn \\
  \addr
 State Key Laboratory of Computer Science \\
  Institute of Software \\
  Chinese Academy of Sciences \\
  Beijing, China
       \AND
       \name Xianyin Zhang  \email xianyinzhang22@m.fudan.edu.cn  \\\addr  School of Data Science \\
  Fudan University \\
  Shanghai, China 
  \AND
  Shiyu Huang \email huangsy1314@163.com\\  \addr  4paradigm Inc. \\
  Beijing, China 
  \AND
  Bingzhen Zhang   \email zhangbz23@mails.tsinghua.edu.cn \\
  \addr
  Department of Industrial Engineering \\
  Tsinghua University \\
  Beijing, China \AND
  Ke Wei  \thanks{Correspondence may be addressed to
  Ke Wei and Shaowei Cai.}  \email kewei@fudan.edu.cn \\
\addr
  School of Data Science \\
  Fudan University \\
  Shanghai, China 
  \AND
  Shaowei Cai \footnotemark[1] \email caisw@ios.a.cn \\
  \addr
  State Key Laboratory of Computer Science \\
  Institute of Software \\
  Chinese Academy of Sciences \\
  Beijing, China \\
  (version 2)
}

\begin{document}

\maketitle

\begin{abstract}
Conflict-Driven Clause Learning (CDCL) is the mainstream framework for solving the Satisfiability problem (SAT), and CDCL solvers typically rely on various heuristics, which have a significant impact on their performance. Modern CDCL solvers, such as MiniSat and Kissat, commonly incorporate several heuristics and select one to use according to simple rules, requiring significant time and expert effort to fine-tune in practice. The pervasion of Large Language Models (LLMs) provides a potential solution to address this issue. However, generating a CDCL solver from scratch is not effective due to the complexity and context volume of SAT solvers. Instead, we propose AutoSAT, a framework that automatically optimizes heuristics in a pre-defined modular search space based on existing CDCL solvers. Unlike existing automated algorithm design approaches focusing on hyperparameter tuning and operator selection, AutoSAT can generate new efficient heuristics.
In this first attempt at optimizing SAT solvers using LLMs, several strategies including the greedy hill climber and $(1+1)$ Evolutionary Algorithm are employed to guide LLMs to search for better heuristics. Experimental results demonstrate that LLMs can generally enhance the performance of CDCL solvers. A realization of AutoSAT outperforms MiniSat on $9$ out of $12$ datasets and even surpasses the state-of-the-art hybrid solver Kissat on $4$ datasets. 

\end{abstract}

\section{Introduction}

The Satisfiability problem (SAT), as the first proven NP-complete problem~\shortcite{npc}, is essential for numerous domains within computer science. Thus SAT solvers have shown significant impacts in various areas, such as software verification, program analysis, electronic design automation, and so on~\shortcite{sat_application}. Among many approaches for SAT, Conflict driven clause learning (CDCL) is the most popular one and has been applied with remarkable success since its introduction~\shortcite{audemard2009predicting,kochemazov2019maple,cadical,kissat}. Another alternative approach involves local search, which is an incomplete method that exhibits a bias towards searching satisfiable assignments and can be competitive in solving random and hard combinatorial instances~\shortcite{LocalSearch,balint2010improving}. Efforts in combining CDCL and local search solvers have led to hybrid solvers~\shortcite{cadical,kissat,cai2022better,audemard2010boosting,habet2002hybrid,letombe2008improvements,mazure1998boosting}, which have recently achieved state-of-the-art (SOTA) results in SAT Competitions~\shortcite{sat2023result}. In this work, we focus on CDCL solvers, which have been commonly applied in SOTA hybrid SAT solvers~\shortcite{cai2022better}. The recent development of CDCL solvers mainly relies on proposing novel heuristics, such as branching heuristics, to boost the performance of the solvers, and these heuristics are usually designed based on intuition and expertise through trail-and-error approaches~\shortcite{kissat,marques2021conflict}.


Designing heuristics often demands significant time and human effort. Moreover, the continuous development of numerous methods makes it challenging to select the most appropriate approach for specific problem instances. Therefore, Hyper Heuristics (HHs)~\shortcite{hyperheuristics} have been studied addressing these challenges~\shortcite{geneticprogramming,surveyedl}. HHs typically select the best-performing heuristics from a predefined set or generate new heuristics by combining existing heuristic components. However, the optimal performance of HHs is limited by the configuration space predefined by human experts~\shortcite{hhs1,surveyedl}.

Fortunately, Large Language Models (LLMs)~\shortcite{LLMs,GPT4,lin2023swiftsage} provide us with the potential to break or at least extend the limitation of predefined configuration space~\shortcite{yang2023large} because the models have been trained with innumerable knowledge. 
On the other hand, recent work has experimentally proved LLMs' ability to conduct programming tasks~\shortcite{nam2024using,starcoder,autogpt,flows,AutoGen,agentverse}, and researchers have contributed much effort in applying LLMs for optimization. 
For example, FunSearch~\shortcite{FunSearch} is paired with LLMs to generate solutions in the form of programs and evolves these solutions towards better programs for the cap set and knapsack problems. ReEvo and EoH \shortcite{reevo,evolution} apply LLMs to generate heuristics for problems such as online bin packing, traveling salesman, and flow shop scheduling. LLaMEA also applies evolutionary algorithms to search metaheuristics automatically for fundamental continuous optimization problems~\shortcite{van2024llamea}. 

Since LLMs are capable of programming correctly in many scenarios, we are wondering whether they can assist us in optimizing heuristics for CDCL solvers. More precisely, we investigate the following two questions: 
\begin{itemize}
    \item Can LLMs generate a more efficient implementation of CDCL solver than the existing human-designed implementation?
    \item Given a SAT solver, can LLM optimize the implementation of specific heuristics functions?
\end{itemize}

In response to the first question, LLMs can generate code with a basic CDCL framework. However, we have observed that due to the limitations in text length, the generated code tends to be too brief and simplified to meet the standards of modern SAT solvers, see Appendix~\ref{motivation question}. Therefore, we consider providing LLMs with the source code of an existing solver, enabling them to modify the partial codes of a solver to enhance the performance. This led us to the second question. Surprisingly, we found that LLMs can identify many potential directions for modifying existing codes. Thus, we propose leveraging LLMs to modify heuristics functions given the original solver code in the context.


Overall, the main contributions of this paper are as follows:
\begin{itemize}
    \item \textbf{AutoSAT, A Competitive and Expandable SAT Solver Optimization Framework.} We introduce AutoSAT, an open-source CDCL solver optimization framework that autonomously adjusts embedded heuristics for datasets. Experimental results show that AutoSAT significantly enhances the performance of CDCL solvers and achieves competitive results compared to the classic solver MiniSat. Moreover, CDCL solvers optimized by AutoSAT surpass the SOTA SAT Solver Kissat on several datasets. Although the current version of AutoSAT is built on a simple CDCL framework, the modular design of AutoSAT allows for the integration of additional techniques to further improve its performance in the future.
    \item \textbf{A Novel Scheme for LLM-Based Optimizers in Complex Scenarios.} Unlike existing approaches that rely on LLMs to generate entire programs from scratch, we design a modular CDCL framework of SAT solvers, enabling LLMs to optimize solvers within a fine-grained search space. This approach provides a more practical and scalable solution for leveraging LLMs in complex and large-scale programming scenarios.
    \item \textbf{Comprehensive Analysis of LLM-based Optimizers and SAT Solvers.} Using the fine-grained space defined by the modular CDCL framework, we experimentally analyze (1) the process by which LLMs optimize SAT solvers and (2) how the generated heuristics enhance the performance of CDCL solvers. These discussions provide valuable insights for future research on using LLM for optimizing solvers, addressing topics such as the budget allocation of calling LLMs and the innovative design of SAT heuristics.
\end{itemize}

\section{Preliminaries and Related Works}
\subsection{Preliminary Definition on SAT}
\label{pre:sat}
Let $V = \{x_1, x_2, \dots, x_n\}$ be a set of Boolean variables, a \textit{literal} is either a variable $x$ or its negation $\neg x$. A \textit{clause} is a disjunction of literals. A \textit{conjunctive normal form} (CNF) formula $F = C_1 \land C_2 \land \dots \land C_m$ is a conjunction of clauses. For simplicity we assume all clauses are non-tautological, in other words, no variable $x$ occurs positively $(x \in C)$ and negatively $(\neg x \in C)$ in the same clause.

A (partial) mapping $\alpha : V \to \{0, 1\}$ is called an \textit{assignment}. If $\alpha$ maps all variables to a boolean value, it is termed a \textit{complete} assignment; otherwise, it is referred to as a \textit{partial} assignment. The value of a variable $x$ under an assignment $\alpha$ is denoted as $\alpha[x]$. An assignment $\alpha$ satisfies a clause if at least one literal evaluates to true under $\alpha$, and satisfies a CNF formula if it satisfies all its clauses. A CNF formula $F$ is satisfiable if there is at least one satisfying assignment. The empty clause $\bot$ is always unsatisfiable and represents a \textit{conflict}. SAT is the problem of deciding whether a given CNF formula is satisfiable.



\subsection{CDCL solver}
The CDCL solver is composed of a propagate-and-learn framework and a guessing part. CDCL solvers do propagate-and-learn eagerly in practice and the implementations do not differ much among different solvers. However, the guessing policy and the search-restart policy are  important for performance and differ across implementations. Due to the need for manual design and the lack of rigorous mathematical proof, these policies are called heuristics in this paper.

There is a long history of research on heuristics in SAT solvers, among which branching heuristics play a crucial role and continue to significantly impact the performance of CDCL solvers. For example, Variable State Independent Decaying Sum (VSIDS)~\shortcite{vsids} is a family of branching heuristics that assists in selecting the most promising variable to assign a value in the Make Decision phase. Another important branching heuristic is Learning-Rate Branching (LRB)~\shortcite{lrb}, which frames branching as an optimization problem that picks a variable to maximize a metric called learning rate.

Restart heuristics is also essential for enhancing the performance of CDCL solvers. It allows the solver to abandon the current search path and backtrack to a specific decision level, but the learned clauses are usually maintained for the following search. Fast restart~\shortcite{ramos2011between} is now widely used. Originally Luby restarts were heavily used because they represent a priori optimal strategy~\shortcite{luby}. However, most implementations switched to Glucose-style restarts~\shortcite{glucose}, which are essentially a requirement for prevailing in the SAT Competition. For an extensive overview of these techniques, see the survey presented in~\shortciteA{biere2015evaluating}.

Rephase is another technique that introduces heuristics into CDCL solvers. Its primary objective is to reset or adjust the current partial assignment, thereby enabling the solver to explore diverse search space. PrecoSAT and PicoSAT~\shortcite{precosat} utilize a Jeroslow-Wang score~\shortcite{jeroslow1990solving} to adjust the saved phases either on all or only on irreundant clauses in regular intervals, following a Luby sequence. StrangeNight~\shortcite{Strangenight} employs a strategy of flipping values with a certain probability that depends on the depth of the assignment. The motivation is to avoid the heavy-tail phenomenon. What is more, negative results of the SAT solver Riss are reported in~\shortciteA{balint2015overview}, and this is the first time that several rephasing heuristics were compared and used. 

\subsection{LLM-based Optimizers}
\label{pre:llm}
Leveraging Large Language Models (LLMs) for optimization has become a growing area of research, particularly following DeepMind's success in applying LLMs to solve complex mathematical and bin-packing problems in FunSearch~\shortcite{FunSearch}. However, FunSearch requires substantial computational resources, making it impractical for general applications. That being said, it demonstrates the significant potential of LLMs in tackling hard optimization problems. This has spurred interest in studying LLM-based approaches across various domains. For example, the Evolving of Heuristics (EoH) work utilized LLMs to solve traveling salesman Problems and scheduling problems~\shortcite{evolution}, and follow-up studies adopted similar techniques for evolving the acquisition functions in Bayesian Optimization~\shortcite{yao2024evolve} and for adversarial attacks~\shortcite{guo2024autoda}. Additionally, the LLaMEA~\shortcite{van2024llamea} explored the domain of black-box continuous optimization by evaluating performance on the well-known BBOB benchmark~\shortcite{bbob}.

As researchers continue to investigate the usage of LLMs for various application domains, developing search policies for LLMs-based algorithm design and implementation has also gained increasing attention~\shortcite{understanding}. Most of the current research utilizes evolutionary algorithms to search for better algorithm designs. The pioneering FunSearch work applies the islands model, a parallel genetic algorithm that requires numerous evaluations, i.e., requesting an LLM-generated answer, for each trial. However, more recent studies have shown that simpler approaches, such as the $(1+1)$ Evolutionary Algorithm (EA), can yield promising results within fewer evaluations. Similar trends were also observed in EoH~\shortcite{evolution}, which employed population-based genetic algorithms. A comprehensive benchmarking study~\shortcite{understanding} has compared various evolutionary search policies for LLMs-based optimization work, revealing distinct behaviors and efficiencies among different approaches.

\subsection{Development in Evolutionary Computation}
Evolutionary computation plays an important role in black-box optimization, and as mentioned above, it has shown a significant impact on LLM-based optimization. Population-based methods, $(1+1)$ EA, and other randomized local search strategies~\shortcite{FunSearch,evolution,reevo} are popular approaches. Note that, $(\mu+\lambda)$~EAs in this paper refer to the algorithms that generate new solutions based on mutation only, in which mutation denotes alternating the values of $\ell$ distinct variables, $\lambda$ is the number of new solutions, i.e., offspring, generated in each iteration, and $\mu$ is the number of solutions that are selected for the next generation. While a wide range of evolutionary techniques has been proposed for various applications, tremendous studies have focused on benchmarking and theoretical analysis of these algorithms.

While experimental benchmarking analysis~\shortcite{COCO,de2024iohexperimenter,bennet2021nevergrad} can provide us guidelines in setting operators and parameters of the algorithms, 
theoretical analysis also plays a critical role in helping us understand the behavior of EAs. The pioneering work of the Schema theorem was one of the earliest efforts in analyzing genetic algorithms~\shortcite{golberg1989genetic}, while level-based methods were later used to analyze the runtime of EAs~\shortcite{wegener2001theoretical}. More recently, the well-known drift analysis has been introduced as a theoretical framework for analyzing EAs~\cite{he2001drift,doerr2012multiplicative,lengler2020drift}. Particularly for the discrete optimization, which is related to the scenario of this work, the study~\cite{doerr2012multiplicative} proved that any linear pseudo-Boolean function can be optimized by $(1+1)$~EA in expected runtime $O(n \log n)$, where $n$ is the dimensionality, i.e., the length of bit strings. For specific functions, the tight bounds of EAs for OneMax, a problem maximizing the number of one-bits in a binary string, have been discussed in detail by~\citeA{witt2013tight}, indicating that $(1+1)$~EA is an optimal mutation-based algorithm for OneMax. Meanwhile, for LeadingOnes,  another classic problem that aims to maximize the number of consecutive one-bits counting from the first position in a binary string, it is proved that EAs achieve an expected optimization time of $O(n^2)$ ~\shortcite{bottcher2010optimal,doerr2018towards}. LeadingOnes is usually used to study how algorithms perform when there is a need for more structured optimization and sequential improvements. 

Since the development of evolutionary computation is a broad topic, we refer readers to the recent surveys~\shortcite{doerr2019theory,back2023evolutionary} for more comprehensive details.

\section{AutoSAT}
\label{sec:autosat}
In this section, we introduce AutoSAT, our proposed automatic LLM-based optimization framework for CDCL solvers. 
AutoSAT aims to enhance SAT solver performance by leveraging LLMs to iteratively search for more effective heuristics.

We first describe the architecture of the modular design CDCL solver and outline the associated search space. Next, we examine two search strategies and present a comparative analysis between the greedy hill climber (GHC) and the $(1+1)$~EA. Lastly, we illustrate effective prompt engineering techniques employed in AutoSAT.

\subsection{Fine-Grained Search Space}
Even though previous studies mentioned in Section~\ref{pre:llm} have successfully addressed similar optimization problems in other domains, none have explored scenarios involving SAT, which require high computational efficiency. SAT solvers usually employ advanced programming techniques to deliver results within minimal cputime. Over decades of development, SOTA SAT solvers, such as Kissat, have integrated numerous heuristics. However, in our preliminary experiments, using LLMs to generate a SAT solver from scratch following the EoH mechanism only produces a basic Davis–Putnam–Logemann–Loveland (DPLL) framework (see Appendix~\ref{sec:eoh-exp}). Even after several iterations, the generated solver tended to include only a single heuristic module for improvement. Moreover, guiding LLMs to design competitive data structures remains particularly challenging, especially compared to to the highly optimized implementations of SOTA SAT solvers that have been refined over decades.
\begin{algorithm2e}[t]
\caption{CDCL Framework}
\label{alg:CDCL}
\textbf{Input:} A CNF $F$ of SAT instance\;
\textbf{Initialization: } decision level $d \leftarrow 0$, current assignment of variables $\mathcal{X} \leftarrow \emptyset$\;

\While{\textbf{True}}{
$\mathcal{X} \leftarrow \textbf{Unit Propagation}(F,\mathcal{X})$\;
\eIf{Conflicts are detected in $\mathcal{X}$}{
\eIf{$d == 0$}{return \textsc{UNSAT}}
{$\mathcal{C}_{conflict}, d_{backtrack} \leftarrow \textbf{Analyze Conflict}(F,\mathcal{X})$ \;
$C_{learned} \leftarrow \textbf{Learn Clause}(\mathcal{C}_{conflict},\mathcal{X})$ \;
$F \leftarrow F \land C_{learned}$\;
$\mathcal{X} \leftarrow \textbf{Backtrack}(\mathcal{X},d_{backtrack})$\;
$d \leftarrow d_{backtrack}$\;
}
}
{
\eIf{All variables are assigned with a value}{return \textsc{SAT}}
{
$d \leftarrow d + 1$\;
$\mathcal{X} \leftarrow \textbf{Make Decision}(\mathcal{X})$
}
}
}
\end{algorithm2e}

To address these challenges, we use a predefined, modular CDCL framework as presented in Alg.~\ref{alg:CDCL}. The CDCL algorithms usually initiate with an empty set of partial assignments (line 2). The Unit Propagation (UP), also called Boolean Constraint Propagation, assigns values to variables that are the only ones that can make a clause satisfied (i.e., all other literals in this clause are false). This operation will repeat until no more UP is possible (line 4). If no conflicts are detected in $\mathcal{X}$ during the Decision Detection phase, the algorithm will select a variable and assign it a value (line 19). This Make Decision step usually follows heuristics. When conflicts are detected (line 5), Analyze Conflict will identify the conflicted clauses (line 9), and a newly learned clause will be derived based on the cause (i.e., the current partial assignment) of conflicts (line 10). Afterward, the algorithm backtracks to an earlier decision level (lines 12-13). Once all variables are assigned and no conflict is detected, the algorithm obtains a satisfied assignment (line 16). On the other hand, detecting conflicts at the decision level $0$ indicates that the given CNF is unsatisfiable (line 7).

Modern SAT solvers are usually built on the CDCL framework, and various heuristics have been continuously proposed to enhance their performance. For instance, \emph{reduce} heuristics works on the tracks of learned clauses (referring to line 10). They control the size of the tracking list, identifying and removing learned clauses. In addition, \emph{bump var heuristics} usually incorporate with the Analyze Conflict step (line 9) and affect the choice of variables in the Make Decision function (line 19). While the order of choosing variables can determine the search path of branching, 
\emph{rephase} heuristics can control the polarity in variales to be selected. Also,
\emph{restart} heuristics may abandon the current search path allowing algorithms to explore possibly more easier search regions.

In this paper, we work on a modular CDCL framework that defines $nine$ functions which are independently implemented: \textbf{restart}, which manages restart heuristics, \textbf{restart condition update}, which adjusts restart conditions, \textbf{restart condition}, which determines when to execute restart, \textbf{reduce}, which handles reduce heuristics, \textbf{reduce condition}, which determines when to reduce, \textbf{rephase}, which manages rephase heuristics, \textbf{rephase condition}, which determines when to rephase, and two functions \textbf{bump var} and \textbf{bump var heuristic} that govern the order of variables being selected during conflict analysis. AutoSAT operates within a search space of heuristics $H: \{h_1, h_2, \ldots, h_9\} \mapsto A$, where each $h_i$ represents a heuristics for a function, and the nine functions together form a customized CDCL Solver $A$. Note that the candidates for each $h_i$ depend on the behavior of LLMs. In practice, we implement this modular CDCL solver based on an open-source EasySAT project~\shortcite{EasySAT}.

\subsection{Search Strategies}
\label{search strategies}
\paragraph{Optimization Problem} Based on the search space defined above, optimizing CDCL solvers can be formulated as a minimization problem: given the search space
$H=\{(h_1, \ldots, h_9)~|~ h_i \in H_i, i=1, \ldots ,9\}$
 and a group of SAT instances $P$,  the objective is to find a configuration $h^*$ such that 
\begin{align}
    h^* \in \underset{h \in H}{\arg \min}  f(h,P).
\end{align}
In this paper, we evaluate $f(h,P)$ by the PAR2 value of $A$ across $P$. In addition,
since there is a one-to-one correspondence between $A$ and $h\in H$, we further denote $f(h,P)$ by $f(A)$ in the sequel. Note that it remains challenging to investigate the size of candidates for each heuristics function $h_i$.

\paragraph{Algorithms}
We examine the \emph{greedy hill climber (GHC)} and \emph{$(1+1)$~ Evolutionary Algorithm (EA)} for our AutoSAT to optimize CDCL solvers. The GHC is an exact algorithm that modifies the heuristic functions going from the first to the last one. It can efficiently achieve immediate improvement during searching, but it can easily get stuck in local optima due to its strict improvement criterion and lack of diversity mechanisms. The $(1+1)$~EA is a classic EA that operates with a single parent solution and generates new solutions by modifying one or multiple randomly chosen heuristic functions. Compared to the GHC, it obtains advantages in exploration. Both algorithms are well suited to our optimization scenario, and many fundamental studies have been conducted for their experimental and theoretical properties~\shortcite{doerr2020benchmarking}. Therefore, we examine both approaches, providing a baseline for future comparisons.


In practice, we show in Alg.~\ref{alg:autosat_greedy} the procedure of AutoSAT embedded with the GHC. AutoSAT starts with a solver in which nine heuristics functions follow the settings in EasySAT~\shortcite{EasySAT}\footnote{We use EasySAT instead of MiniSat or Kissat due to its light volume. More precisely, EasySAT requires only $5k$ tokens when inputing into LLMs, in contrast to MiniSat's $25k$ tokens and Kissat's $250k$ tokens, which makes EasySAT more manageable for LLMs to understand and revise. Furthermore,  Kissat is lack of modularity in the codebase, resulting in a tangled structure that poses significant challenges for modifications by LLMs.}. In each iteration, AutoSAT modifies a chosen heuristic function to generate a new solver, and the nine heuristic functions are selected in order, starting from the first to the last one (line 4). After determining the choice of the heuristic function, AutoSAT requests LLMs to provide alternative implementations for the chosen heuristics functions (lines 5-6). In each iteration, based on the LLM-generated response, AutoSAT generates and evaluates the new solver $A'$ (lines 7). If $A'$ outperforms the current best solver $A$, AutoSAT will greedily update the heuristics accordingly, incorporating the best implementation so far (lines 8-10).

\begin{algorithm2e}[t]
\caption{AutoSAT Framework with the Greedy Hill Climber (AutoSAT\_GHC)}
\label{alg:autosat_greedy}
\textbf{Input:} A group of SAT instances $P$, a modular CDCL formed by nine independent functions $\{h_1,h_2,\ldots, h_9\}$, a prompt template based on the current best $A$ and indexes $M \leftarrow \emptyset$ of functions to be modified\;
\textbf{Initialization: } Assign a predefined $h \leftarrow \{h_1,h_2,\ldots, h_9\}$, and initialize a solver $A$ based on $h$, $f^* \leftarrow$ evaluate the performance of of $A$\;

\For{$i \in \{0,\ldots,\mathcal{B}\}$}{
    Get function index $m \leftarrow  i \mod 9 $\;
    Form a new prompt based on $A$ and $m$, and request new heuritics functions $\{h'_{m}\}$ from LLMs\;
    $h' \leftarrow $ replacing the functions in $h$ by the corresponding functions in $\{h'_{m}\}$\;
    Obtain a new solver $A'$ based on $h'$, and $f(A') \leftarrow$ evaluate the performance of $A'$\;
    \If{$f(A') \leq f^*$}{$A \leftarrow A'$ and $h \leftarrow h'$\; $f^* \leftarrow f(A')$}
}   
\end{algorithm2e}

Although the GHC can intuitively make deterministic achieving immediate improvement in optimizing the SAT solvers, it can be trapped in a local optimum without any diversity mechanism. In addition, the design of modifying heuristic functions one by one is built under the assumption that different heuristic functions contribute to the performance of the SAT solvers independently, which is likely not to be the truth. For example, three heuristics functions \textbf{restart condition}, \textbf{restart condition update}, and \textbf{restart} can not work independently. Therefore, we also examine the performance of $(1+1)$~EA for AutoSAT.

In Alg.~\ref{alg:autosat}, AutoSAT also starts with an initial solver following the settings in EasySAT~\shortcite{EasySAT}. It then iteratively requests LLMs to provide alternative implementations for $\ell$ distinct heuristics functions (lines 4-6). Instead of modifying a deterministically chosen heuristic function  (line 4 in Algorithm~\ref{alg:autosat_greedy}), the $(1+1)$~EA approach selects $\ell$ distinct heuristic function to mutate (line 5), and $\ell$ follows a static binomial distribution (line 4). In this way, AutoSAT is allowed to generate a new solver $A'$ by modifying one or multiple heuristic functions(lines 7-8). If $A'$ outperforms the current best solver $A$, AutoSAT will also greedily update the prompt accordingly, incorporating the best implementation so far (lines 9-11).


It is worth noting that the fine-grained search space of AutoSAT allows us to leverage existing research in discrete optimization to design our search strategies. Recalling the problem definition of optimizing CDCL solvers, if LLMs can suggest the optimal candidates of each function $h_i$ when asked about optimizing heuristics, this problem can align with a pseudo-Boolean optimization problem. Based on this assumption and considering that the performance of a heuristic function $h \in H$ may be interdependent with others, searching for the optimal AutoSAT solver can be approached as a sequence problem. Drawing from research on the classic LeadingOnes problem~\shortcite{bottcher2010optimal}, we set our budget $B \approx 0.6n^2, n = 9$, following the established benchmarking and theoretical insights into LeadingOnes~\shortcite{doerr2018towards,ye2020benchmarking}. While our assumptions may not strictly reflect objective truth, the AutoSAT approach establishes a valuable framework linking foundational research and practical applications in LLM-based optimization.

\begin{algorithm2e}[t]
\caption{AutoSAT Framework with $(1+1)$~EA (AutoSAT\_EA)}
\label{alg:autosat}
\textbf{Input:} A group of SAT instances $P$, a modular CDCL formed by nine independent functions $\{h_1,h_2,\ldots, h_9\}$, a prompt template based on the current best $A$ and indexes $M \leftarrow \emptyset$ of functions to be modified\;
\textbf{Initialization: } Assign a predefined $h \leftarrow \{h_1,h_2,\ldots, h_9\}$, and initialize a solver $A$ based on $h$, $f^* \leftarrow$ evaluate the performance of of $A$\;

\For{$i \in \{0,\ldots,\mathcal{B}\}$}{
    Sample a value $\ell \sim \text{Bin}(9,\frac{1}{9})$\;
    chosen $\ell$ distinct values $M \leftarrow \{m_0, \ldots, m_\ell\}$ from $\{1,\ldots,9\}$ uniformly at random\;
    Form a new prompt based on $A$ and $M$, and request new heuritics functions $\{h'_{m_0}, \ldots, h'_{m_l}\}$ from LLMs\;
    $h' \leftarrow $ replacing the functions in $h$ by the corresponding functions in $\{h'_{m_0}, \ldots, h'_{m_l}\}$\;
    Obtain a new solver $A'$ based on $h'$, and $f(A') \leftarrow$ evaluate the performance of $A'$\;
    \If{$f(A') \leq f^*$}{$A \leftarrow A'$ and $h \leftarrow h'$\; $f^* \leftarrow f(A')$}
}   
\end{algorithm2e}

\subsection{Prompt Engineering}
Prompt Engineering is essential for the effective use of LLMs, ensuring that LLMs operate within the intended context and leverage ongoing outputs, e.g., new heuristics implementation generated by AutoSAT. We follow the instructions in OpenAI framework docs~\cite{openai_docs} and make appropriate embellishments and modifications based on actual usage. Our prompt template obeys the following format:

\begin{itemize}
    \item Define the \textbf{Role} of an agent as a solver expert who needs to assess and improve the heuristics in an SAT solver.
    \item Clearly state the \textbf{Goal}, such as providing optimization suggestions, writing code, or feedback.
    \item 
    Enhance the agents' capabilities by inserting optional \textbf{Tips} that guide them to avoid common mistakes during code generation. Additionally, through this flexible interface, agents can effectively utilize external codes and results and can be instructed to specify the types of modification directions such as changing parameters, modifying heuristics, or adding new heuristics.
    \item Total SAT solver code is appended at the end of each prompt to ensure all agents are in the same context.
\end{itemize}
Detailed prompt templates in AutoSAT are available in appendix~\ref{sec:prompt}. 




\section{Experimental Results} 
In this section, we present the experimental results of AutoSAT across $12$ datasets, addressing the topics of improving the baseline solver, comparisons against the SOTA solvers, and the convergence process of searching for better solvers.

\subsection{Settings}
\paragraph{Environment} SAT solvers are implemented in C++, and the interface to LLMs is implemented by Python. We compile the generated solver using g++ 9.4.0, and all experiments are conducted on three servers, each equipped with \textit{AMD EPYC 7763 64-core Processors}. The budget, denoted as $\mathcal{B}$, referring to the times requesting new heuristics from LLMs, is $60$ (larger than $0.6 * 9^2$, see Section~\ref{search strategies}) for testing AutoSAT on each group of instances. The timeout bound $\mathcal{T}$ during experiments is $5000$s for each SAT instance. Owing to the superior performance of AutoSAT when utilizing GPT-4o\footnote{We utilize the GPT-4o-2024-08-06 as default in this paper.}, we only report results obtained from it in this paper.

\paragraph{Benchmarks} We selected $12$ datasets to explore the capabilities of AutoSAT, consisting of $7$ datasets from SAT Competition 2023~\shortcite{sat2023}, and $5$ datasets generated by Picat. 
For the selection of SAT competition datasets, we filter out families with fewer than twenty instances, and this results in a total of $7$ families of instances: argumentation, cryptography-ascon, set-covering-with-pairs (SCP), register-allocation, social-golfer, hashtable-safety, and profitable-robust-product (PRP). \footnote{In SCP, we collect $20$ instances from SAT Competitions 2023, and $13$ instances of the same group from SAT Competitions 2022 and 2018.}
The remaining $5$ datasets are manually generated by the Picat tool, including CoinsGrid, LangFord, KnightTour,  MineSweeper and Zamkeller, which can formulate constrained satisfied problems into CNF formulas~\shortcite{picat}. We summarize sources and characteristics of all benchmark sets in Tab.~\ref{tab:bench}, more details about how to generate are shown in Appendix~\ref{generation}.

\begin{table*}[t]
\belowrulesep=0pt
\aboverulesep=0pt
\begin{center}
\caption{\textbf{Dataset information}. We summarize datasets tested in this paper from SAT Competition 2023 (SC 2023 for short) and Picat (a language tool for problem generation). The size of each dataset and statistics on the number of variables and clauses are provided here, illustrating the diversity in problem complexity. More details are shown in Appendix \ref{dataset_details}.}
\vskip 0.04in
\resizebox{\textwidth}{!}{
\begin{tabular}{lcc|cc|cc}
\toprule
Dataset            & Size & Source & variables mean & variables std & clauses mean & clauses std \\
\midrule
cryptography-ascon           & 20 & SC 2023 & 146,636 & 15,010 & 342,940 & 37,315 \\
register-allocation       & 20  & SC 2023 & 381 & 193 & 5,813 & 5,622 \\
social-golfer          & 20 & SC 2023 & 15,540 & 12,157 & 131,517 & 79,751\\
hashtable-safety              & 20  & SC 2023 & 11,712,548 & 4,874,397 & 53,644,509 & 22,032,387 \\
PRP    & 20 &SC 2023  & 3,040,711 & 3,281,321 & 18,338,492 & 19,807,576 \\
argumentation             & 20  & SC 2023 & 962 & 190 & 27,960 & 23,034 \\ 

SCP      & 33  & SC 2023 et al. & 682 & 164 & 26,969 & 10,830 \\
\midrule
CoinsGrid             & 98  & Picat & 3,040,711 & 3,281,321 & 18,338,492 & 19,807,576 \\
MineSweeper              & 88  & Picat & 618,801 & 531,657 & 9,065,224 & 7,909,645  \\
KnightTour           & 56  & Picat & 223,697 & 313,704 & 10,692,883 & 17,482,123  \\
LangFord              & 134  & Picat & 272,117 & 229,927 & 2,376,022 & 2,112,183 \\
Zamkeller              & 80  & Picat & 24,592 & 22,190 & 310,804 & 335,102 \\

\bottomrule
\label{tab:bench}
\end{tabular}
}
\end{center}
\vskip -0.08in
\end{table*}

\begin{table*}[t]
\label{PAR2result}
\belowrulesep=0pt
\aboverulesep=0pt
\begin{center}
\caption{\textbf{PAR2 over different dataset}. This table presents PAR2 (lower is better) of the tested solvers across various datasets, followed by the numbers in brackets presenting the number of instances solved within the time limit (more is better). Apart from the results of the tested solver, i.e., EasySAT, AutoSAT\_GHC, AutoSAT\_EA, MiniSat, and Kisssat, we also list the percentage of PAR2 improvements comparing the better one of AutoSAT\_GHC and AutoSAT\_EA to EasySAT in the column ``Improvement''.}
\vskip 0.04in
\resizebox{\textwidth}{!}{
\begin{tabular}{lcccc|cc}
\toprule
Dataset & EasySAT & AutoSAT\_GHC & AutoSAT\_EA & Improvement & MiniSat & Kissat \\
\midrule

cryptography-ascon & 503 (20) & 215 (20) & 275 (20) & 57.26\% & 6,173 (8) & \textbf{146 (20)} \\
register-allocation & 10,000 (0) & 10,000 (0) & \textbf{1,178 (18)} & 88.22\% & 8,808 (3) & 8,554 (5) \\
social-golfer & 10,000 (0) & \textbf{7,265 (6)} & 7,297 (6) & 27.35\% & 9,009 (2) & 7,536 (5) \\
hashtable-safety & 1,041 (20) & 684 (20) & 674 (20) & 35.25\% & \textbf{229 (20)} & 572 (20) \\
PRP & 10,000 (0) & 5,204 (10) & 5,044 (10) & 49.56\% & 8,537 (3) & \textbf{5,015 (10)} \\
argumentation & 10,000 (0) & 9,535 (1) & 9,077 (2) &  9.92\% & 5,240 (12) & \textbf{3,330 (14)} \\
SCP  & 10,000 (0) & 8,992 (5) & 7,466 (10) & 25.34\% & 6,137 (13) & \textbf{76 (33)}	 \\
\midrule
CoinsGrid & 6,855 (32) & 6,372 (37) & 5,593 (46) & 18.41\% & 7,463 (26) & \textbf{4,220 (61)} \\
MineSweeper & 3,886 (63) & 1,608 (83) & 1649 (83) & 58.62\% & \textbf{9 (88)} & 146 (88) \\
KnightTour & 8,952 (6) & 8,633 (8) & \textbf{8,248 (10)} & 7.86\% & 8,602 (8) & 8,754 (7) \\
LangFord & 8,856 (17) & 7,420 (33) & 7,421 (33) & 16,21\% & 8,567 (17) & \textbf{5,567 (65) }\\
Zamkeller & 3,668 (54) & \textbf{2,053 (66)} & 2,823 (62) & 44.03\% & 8,569 (13) & 2,230 (65) \\

\bottomrule
\label{tab:par2}
\end{tabular}

}
\end{center}
\vskip -0.08in
\end{table*}

\paragraph{Performance Metric}
When evaluating a SAT solver, we consider in this paper two specific metrics: (1) The number of SAT instances solved within the given timeout bound, and (2) The Penalized Average Runtime with a factor of 2 score (PAR-2). PAR-2 represents the average run time of a SAT solver for a set of instances, and for the instances where the solver can not return a result within $\mathcal{T}$, the run time is penalized as $2\mathcal{T}$. More detailed explanations of the PAR-2 metric can be found in Appendix~\ref{par2}. Both metrics have been commonly applied in SAT Competitions, and we use PAR-2 to evaluate the performance $f(A)$ of a solver $A$ in Alg.~\ref{alg:autosat_greedy} and Alg.~\ref{alg:autosat}.

\subsection{Results}
\begin{figure*}[t] 
    \centering
    \begin{minipage}{0.5\textwidth}
        \includegraphics[width=\textwidth]{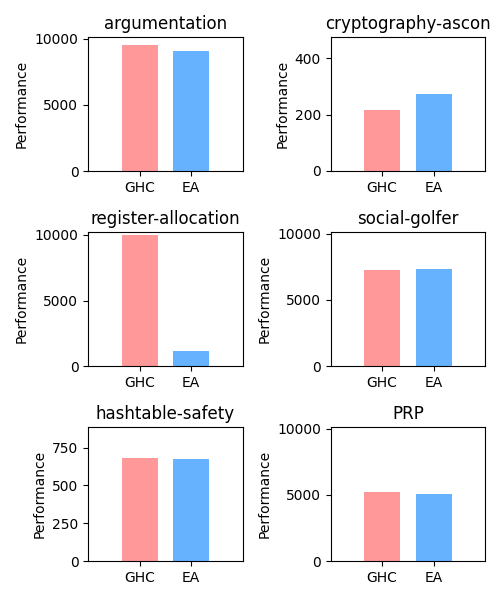}
    \end{minipage}\hfill
    \begin{minipage}{0.5\textwidth}
        \includegraphics[width=\textwidth]{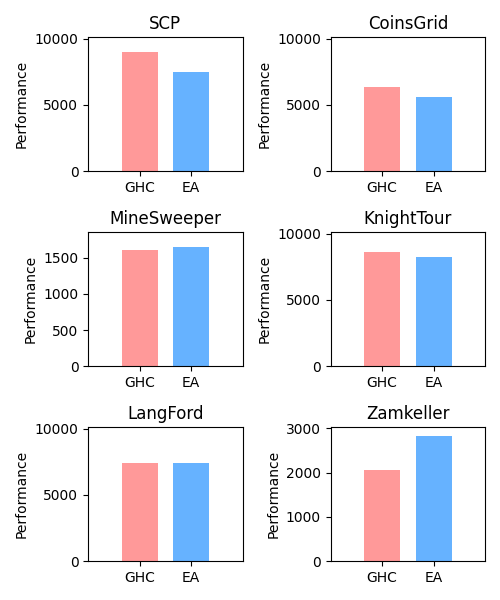}
    \end{minipage}

\caption{\textbf{PAR2 obtained by Different Search Strategies.}  This figure visualizes the comparison between AutoSAT with the greedy hill climber (GHC for short) and the $(1+1)$~EA (EA for short) regarding the PAR2 (lower is better). Overall, the EA outperforms the GHC, but on certain datasets, the GHC even achieves SOTA results.}
\label{fig:strategy}
\end{figure*}

\paragraph{Comparing Search Strategies} 
We show in Fig.~\ref{fig:strategy} the PAR2 of AutoSAT using the GHC (AutoSAT\_GHC) and the $(1+1)$~EA (AutoSAT\_EA) across 12 datasets. Overall, AutoSAT\_EA outperforms AutoSAT\_GHC. However, on certain datasets, AutoSAT\_GHC demonstrates superior performance and can even achieve SOTA results. This may be related to the characteristics of the heuristics functions that perform better after modification for particular datasets. Recall that we set maximum times $\mathcal{B}$ for requesting new heuristics, following studies on LeadingOnes, where $(1+1)$~EA has been shown to outperform the GHC. Our results indicate, however, that the problem of finding the optimal combination of heuristics for CDCL is not directly comparable to LeadingOnes. For example, the candidate implementation for each heuristic function is not binary, and the interdependencies among heuristic functions highlight a need for further investigation. This suggests the necessity of exploring more refined strategies for specific LLM-based optimization scenarios. Although identifying the best strategy for LLM-based optimization is beyond the scope of this initial study on automated SAT solver, our findings provide a foundation for future research into tailored strategies for more customized and advanced scenarios.

\paragraph{Competitive Results of AutoSAT} Since the initialization of our AutoSAT follows the setting of EasySAT, we use EasySAT~\shortcite{EasySAT} as the baseline for the first comparison in this section. As shown in Tab.~\ref{tab:par2}, we can observe that LLMs can obtain significant improvement regarding PAR-2 compared to EasySAT across all datasets. Meanwhile, AutoSAT can solve more instances than EasySAT for $10$ out of $12$ tested instances. These observations clearly indicate that \emph{LLMs is capable of designing and implementing better solvers compared to the provided baseline.}

Furthermore, we compare the other two SOTA SAT solvers, MiniSat~\shortcite{MiniSat} and Kissat~\shortcite{kissat}. Both solvers are widely used in various SAT-related applications, and as well-developed projects, they are highly optimized for efficiency. Notably, the current version of Kissat\footnote{We are utilizing the version of SAT Competition 2023 in https://github.com/arminbiere/kissat.} contains various heuristics, enabling it to select specific strategies based on predefined rules. More details about the two solvers are available in Appendix~\ref{Solver Description}. Our comparative focus is primarily on MiniSat, as it represents a classic CDCL solver with several additional strategies beyond EasySAT. In contrast, Kissat, as a representative of cutting-edge solvers, incorporates a multitude of advanced techniques, far surpassing EasySAT. It sets an exceedingly high bar for CDCL solvers based on EasySAT to meet. 

\begin{figure*}[t] 
    \centering
    \begin{minipage}{0.5\textwidth}
        \includegraphics[width=\textwidth]{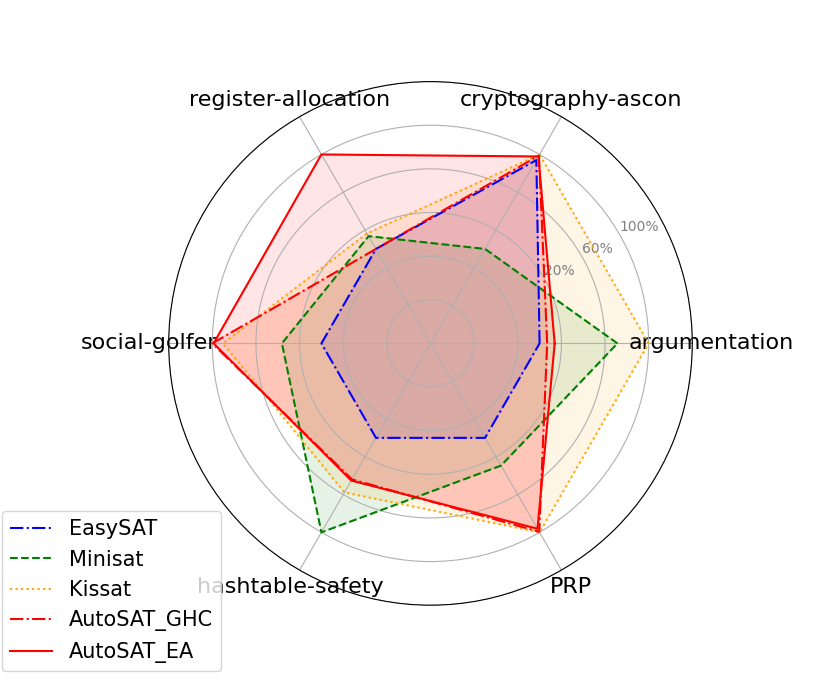}
        \label{fig:left_image}
    \end{minipage}\hfill
    \begin{minipage}{0.5\textwidth}
        \includegraphics[width=\textwidth]{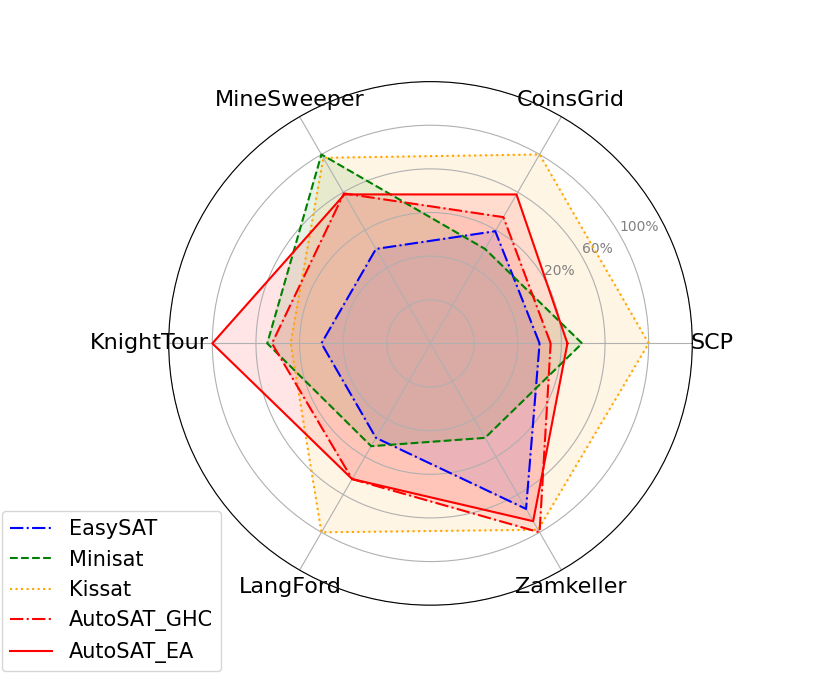}
        \label{fig:right_image}
    \end{minipage}
\caption{\textbf{PAR2 over Different Datasets}. This figure visualizes the comparison of solvers regarding the PAR2 values in table~\ref{PAR2result}. In practice, the plotted values are normalized for each dataset by $1 - \frac{curr - min}{2*(max - min)} $, where $curr$ is the PAR2 obtained by the corresponding solver, and $min$ and $max$ correspond to the minimum and maximum PAR2 values obtained by all the tested solvers, respectively. Larger shadow areas indicate better performance.}
\label{fig:compare}
\end{figure*}

Even though EasySAT employs a simple CDCL framework in contrast to these two complex and high-performance solvers, after iterations of modifications by LLMs, AutoSAT can achieve promising results that are comparable to MiniSat and Kissat. More precisely, AutoSAT outperforms MiniSat on $9$ out of $12$ datasets, and it outperforms Kissat on $4$ datasets. As illustrated in Fig.~\ref{fig:compare}. When compared to Kissat, AutoSAT can still obtain advantages for several datasets. 


\paragraph{Fine-grained search space can benefit LLMs for better solvers.} Unlike other works, such as EoH utilizing LLMs to generate optimization solvers from scratch~\shortcite{evolution}, AutoSAT searches for improved solvers by working on a fine-grained $9$ dimensional space. This approach \emph{allows AutoSAT to optimize SAT solvers, which are usually with long text codes, and to search for executable and better solvers more efficiently.} We have conducted experiments following the approach of EoH~\shortcite{evolution}, but the obtained answers mostly followed a simple CDCL framework and are not comparable to MiniSat and Kissat. However, as shown in the results above, our AutoSAT is comparable to MiniSat and Kissat.

It is also worth mentioning that the fine-grained search space can serve as a bridge, transferring existing benchmarking and theoretical research on discrete optimization to LLM-based searching methods. For example, we leverage the theoretical study on LeadingOnes~\shortcite{doerr2018towards} to set up our budget by $\mathcal{B} = 60$ (larger than $0.6 * 9^2$) (see Section~\ref{search strategies}). 

\begin{figure}[t]
    \centering
    \includegraphics[width=1\textwidth]{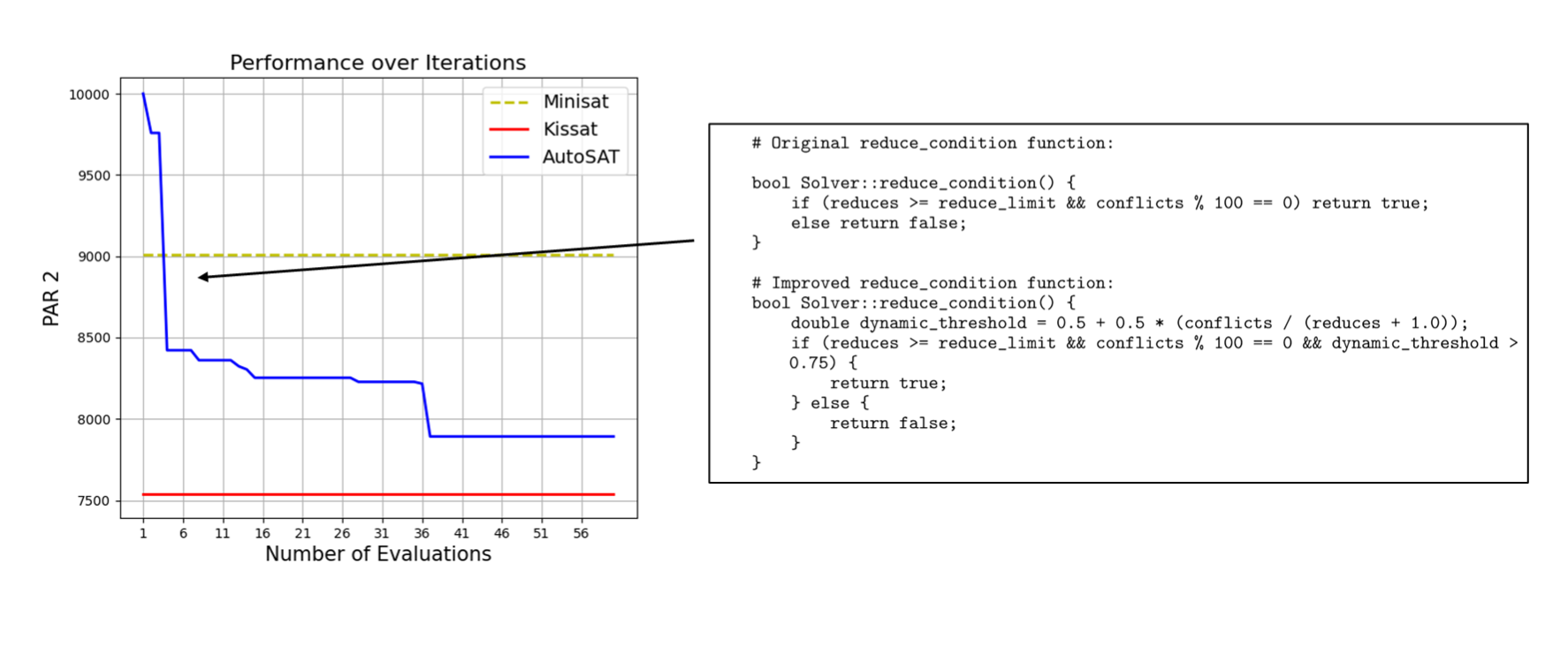}
        \caption{\textbf{PAR2 during searching process in register-allocation dataset}. This figure illustrates an example of AutoSAT searching for better heuristics.
        As shown in the figure, ten modifications result in improvements of PAR2 along the optimization process.
        We highlight on the right a modification of the $reduce\_condition\_functin$ heuristics, which happens at the forth evaluation.
        }
    \label{fig:converge}

\end{figure}

\paragraph{Useful Searching History} Taking the experiment on register-allocation dataset as an example, we illustrate the search process of AutoSAT with $(1+1)$~EA in Fig.~\ref{fig:converge}. Benefiting from the fine-grained search, we can easily spot that, within the $\mathcal{B} = 60$ answers given by LLMs, AutoSAT achieved improvements in $10$ evaluations, including:
\begin{itemize}
    \item updating $reduce$ heuristic function $2$ times;
    \item updating $restart\_condition$ heuristic function $3$ times;
    \item  updating $rephase\_condition$ heuristic function $3$ times;
    \item updating $rephase$ heuristic function $2$ times;
    \item updating $heuristics\_bump\_var$ heuristic function $1$ times; 
    \item updating $reduce\_condition$ heuristic function $2$ times.
\end{itemize}
 Note that our approach of using an EA to optimize modularized solver supports modifying more than one heuristic at once.  Precise updated information on each heuristic function can be useful in the future for adaptive searching strategies since it reflects which heuristic functions obtain significant impact on the performance of SAT solvers for specific instances. In addition, compared to other work generating codes from scratch, this final obtained solver can provide clearer guidelines and more useful knowledge for SAT researchers.

As shown in Fig.~\ref{fig:converge}, the modification of $reduce\_condition$  heuristic functions results in significant improvements during the search process, and we use it as examples to show what we can learn from LLMs.

\begin{lstlisting}[framerule=0pt]
# The variable reduce_limit is a static threshold.
# The variable conflicts indicates the total times that a conflict occurs.
# The variable reduces indicates the times that a conflict occurred since the last round of reduce.

# Original reduce_condition function:

bool Solver::reduce_condition() {
    if (reduces >= reduce_limit && conflicts % 100 == 0) return true;
    else return false;
}

# Improved reduce_condition function:
bool Solver::reduce_condition() {
    double dynamic_threshold = 0.5 + 0.5 * (conflicts / (reduces + 1.0));
    if (reduces >= reduce_limit && conflicts % 100 == 0 && dynamic_threshold > 0.75) {
        return true;
    } else {
        return false;
    }
}


\end{lstlisting}
The original $reduce\_condition$ function is based on simple conditions if the times that a conflict occurred since the last round of reduce, i.e., variable $reduce$, has reached a specified threshold, i.e., variable $reduce\_limit$, and whether the number of conflicts is a multiple of $100$. If both conditions are met, the function returns true, indicating that a reduce should occur. In contrast, the improved version introduces an additional dynamic threshold, as determined by the formula $0.5 + 0.5 * (conflicts / (reduces + 1.0))$, in which variable $conflicts$ indicates the total times that a conflict occurs. As the time that a conflict occurs increases, this dynamic threshold will be met with a smaller value of $reduce$, adding nuance to the condition of reduce by more accurately reflecting the solver's current state. By integrating this dynamically shifting threshold, the solver gains a clear enhancement in solving the register-allocation dataset. 


\section{Conclusion and Discussion}
\label{sec:conclusion}
We introduced AutoSAT, an innovative framework that leverages LLMs to optimize the heuristic functions of modular designed CDCL solvers. Unlike traditional LLM-based optimization approaches, which typically focus on generating code from scratch, AutoSAT integrates expert knowledge to guide LLMs in refining specific heuristics functions, thereby producing competitive SAT solvers automatically. In practice, we apply AutoSAT to a modular CDCL solver including nine heuristic functions, using GHC and  $(1+1)$~EA to iteratively select and optimize heuristic functions through alternative implementation generated by LLMs.

With its default setting, i.e., the modular design of the initial CDCL solver and $(1+1)$ EA, AutoSAT itself serves as a self-adjusting, automated SAT solver that delivers competitive results compared to the well-known solvers such as MiniSat and Kissat. Specifically, AutoSAT outperformed MiniSat on $9$ out of $12$ tested datasets and achieved the best performance on $4$ datasets. 

This work presents a novel paradigm for LLMs-based optimization by focusing on a fine-grained, pre-defined search space with a modular CDCL implementation. AutoSAT demonstrates clear potential for future improvement. Furthermore, it offers opportunities to deepen our understanding of the LLM-based optimization process and the resulting solvers. As discussed, adaptive rephase heuristics can be particularly effective in solving register-allocation SAT instances.

Following the AutoSAT framework, we aim to incorporate more advanced modules into our initialized CDCL solvers to further improve its performance. In addition, we plan to explore self-adaptive search strategies to enhance solver efficiency. A deeper investigation into the behavior of LLMs will also be crucial for understanding and improving the performance of LLM-based solvers. To this end, we will investigate how and why LLMs modify heuristics, with the goal of establishing a convincing guideline for LLM-based optimization frameworks.

\bibliography{neurips_2024}
\bibliographystyle{theapa}

\newpage
\appendix

\section{Solver Description}
\label{Solver Description}

AutoSAT works on modular designed CDCL solvers such that it can perform optimization on a fine-grained search space. While we design and implement the modular SAT solver with nine independent heuristic functions, the initial settings of these functions follow the classic EasySAT solver\footnote{https://github.com/shaowei-cai-group/EasySAT}. EasySAT is a CDCL solver embedded with the following techniques beyond the basic framework provided in Algorithm~\ref{alg:CDCL}:
\begin{itemize}
    \item \textbf{Branching Heuristics}: EasySAT implements branching heuristics named Variable State Independent Decaying Sum (VSIDS), which prioritizes variables for decision-making based on their involvement in recent conflicts. This heuristic helps the solver focus on the most relevant parts of the search space.
    \item \textbf{Restart and Rephase}: EasySAT implements restart strategies to escape from difficult regions of the search space, and rephase strategies to periodically change the phase of variables, which can lead to different search paths and potentially faster convergence.
    \item \textbf{Clause Reduction}: To manage memory and focus on useful clauses, EasySAT periodically removes clauses that are deemed less likely to contribute to finding a solution.
\end{itemize}

Compared with EasySAT, MiniSat implements some more techniques, including:
\begin{itemize}
    \item \textbf{Preprocessing and Simplification}: MiniSat applies various preprocessing techniques to simplify the formula before running the full SAT-solving process. 
    \item \textbf{Data Structure Optimizations}: MiniSat optimizes data structure, significantly contributing to its efficiency and speed in solving SAT. These optimizations are essential for handling large-scale instances and ensuring the solver can process millions of clauses and variables efficiently. 
\end{itemize}

Compared with EasySAT and MiniSat, Kissat implements many more advanced techniques that significantly improve its performance, achieving SOTA in SAT Competitions. We list some key improvements of them, including:
\begin{itemize}
    \item \textbf{Clause Management and Database Reduction}: Kissat introduces more sophisticated methods for managing learned clauses, including glue clause recycling, which measures the quality of learned clauses and removes less useful ones. This approach ensures that the solver keeps a smaller, more relevant set of clauses.
    
    \item \textbf{Enhanced Restart Strategies}: Kissat uses tiered restart strategies, where the solver decides when to restart based on multiple levels of heuristics. This approach improves the balance between local search and global search, leading to faster convergence on difficult problems.
    
    \item \textbf{Data Structure Optimization}: Kissat further optimizes memory usage and data access with highly efficient bit-level compression and cache-friendly designs, which reduce the overhead associated with clause storage and propagation.
    
    \item \textbf{In-Search Simplification}: Kissat continuously applies in-search simplification techniques, including variable and clause elimination, to dynamically reduce the complexity of the problem as the search progresses. This ensures that the solver remains efficient throughout the entire solving process.
    
    \item \textbf{Heuristic Enhancements}: Kissat introduces hybrid and dynamic heuristics for variable selection. For example, combining VSIDS and  Conflict History Based (CHB) heuristics, which adapt to different phases of the solving process to enhance decision-making.
\end{itemize}

\section{PAR-2}
\label{par2}
The Penalized Average Runtime with a factor of 2 score (PAR-2)  is utilized to evaluate the performance of SAT solvers over a benchmark set of instances. It is particularly useful in scenarios where solvers may not return a result for instances within a given timeout bound. The PAR-2 score penalizes such instances by assigning a penalty of twice the timeout bound $\mathcal{T}$.

Consider a benchmark set of $n$ instances. Let $t_i$ be the runtime of the SAT solver on instance $i$. The PAR-2 score is defined as:

\[
\text{PAR-2} = \frac{1}{n} \sum_{i=1}^{n} \tau_i
\]

where: 

\[
\tau_i = 
\begin{cases} 
t_i, & \text{if } t_i \leq \mathcal{T} \\
2\mathcal{T}, & \text{if } t_i > \mathcal{T} \text{ or the solver fails to return a result}
\end{cases}
\]

$\mathcal{T}$ is the predefined timeout bound. Here we give an example to illustrate the computation of the PAR-2 score, here is the example. \\
\textbf{Example}:\\
Assume a benchmark set with 3 instances and a timeout bound $\mathcal{T} = 100$ seconds. The runtimes (in seconds) for the instances are as follows: Instance 1: $t_1 = 80$, Instance 2: $t_2 = 120$, Instance 3: Solver fails to return a result

For Instance 1, since $t_1 \leq \mathcal{B}$, we have $\tau_1 = 80$. 

For Instance 2, since $t_2 > \mathcal{B}$, we have $\tau_2 = 200$ (penalized).

For Instance 3, since the solver fails, $\tau_3 = 200$ (penalized).

The PAR-2 score is calculated as:

\[
\text{PAR-2} = \frac{1}{3} (80 + 200 + 200) = \frac{480}{3} = 160
\]

\clearpage
\section{Dataset}
\label{dataset_details}

\subsection{Dataset Description}
Here we provide brief information about the benchmark datasets including $7$ datasets selected from SAT Competition 2023 and $5$ newly generated datasets. 
\begin{itemize}
    \item \textbf{Argumentation problem} involves finding acceptable sets of arguments in a directed graph where attacks between arguments are represented by edges.
    
    \item  \textbf{Social Golfer problem} is a combinatorial problem that aims to assign golfers to groups over several weeks, ensuring no two golfers play in the same group more than once.
    
    \item  \textbf{Hashtable Safety problem} focuses on verifying the correctness of operations in a hash table to avoid collisions and ensure the integrity of the structure.

    \item \textbf{Register allocation problem} is a problem that arises in compiler optimization, where the goal is to assign a limited number of CPU registers to variables in a program. 

    \item \textbf{Profitable-Robust-Production problem (PRP)} focuses on finding a robust production plan that remains profitable under uncertain or fluctuating conditions. 
    
    \item \textbf{Set Covering with Pairs problem (SCP)} is a variant of the classic Set Covering Problem, which is to find a minimum subset of sets from a given collection, such that the union of these sets covers all elements of a given universe. The additional constraint is that certain sets cannot be chosen together.
    
    \item \textbf{CoinsGrid problem} is derived from Tony Hurlimann’s coin puzzle, focusing on arranging coins on a grid with specific row, column, and distance constraints.

    \item \textbf{MineSweeper problem} is derived from the classic MineSweeper game, where the objective is to determine the placement of hidden mines on a grid based on numerical clues. 
    
    \item \textbf{LangFord problem} is a combinatorial mathematics problem that involves finding a specific permutation of the sequence $1, 1, 2, 2, ..., n, n$ where the two copies of each number $k$ are exactly $k$ units apart.
    
    \item \textbf{KnightTour problem} aims to find a path for a knight on a chessboard that visits every square exactly once, with possible extensions to different board sizes and types.
    
    \item \textbf{Zamkeller problem} involves finding a permutation of integers from $1$ to $n$ that maximizes the number of differential alternations in subsequences divisible by integers from $1$ to $k$, where $(1 < k < n)$.
  
\end{itemize}

\subsection{Problem Generation}
\label{generation}
In this section, we provide the details of generating group of instances using Picat~\shortcite{picat}. Specifically, we adopt scalable problems referring to the settings in Chapters 2 and 3 in the book~\shortcite{picat}. Sampling these instances requires specific parameter settings, and we conducted grid sampling within the parameter space presented below. We define a parameter space $\Theta = \{\theta_1, \theta_2, \ldots\}, \theta^{L}_i \le \theta_i \le \theta^{U}_i$, ensuring that three baseline solvers including EasySAT, MiniSat and Kissat, can obtain a solution within proper cputime range, i.e., [$1$s, $5000$s]. In addition, we apply a space $\Theta'$ to generate our dataset by enlarging the upper bound of $\Theta$, e.g., $\theta'^U_i = \theta'^U_i * 1.2$, such that we can test whether AutoSAT can solve the instances where the baseline solvers can not solve.


\begin{itemize}
    \item \textbf{CoinsGrid problem} \\
    Parameter $\Theta$: $\{n, c\}$ \\
    Parameter Space: $ [16, 157] \times [45, 400]$ \\
    Notes: An $n \times n$ grid with $c$ coins. Each row and column contains exactly $c$ coins.
    \item \textbf{MineSweeper problem} \\ 
    Parameter $\Theta$: $\{m, n, k, p\}$	\\
    Parameter Space: $[500, 1,600] \times [400, 3200] \times [72,689, 1,572,118] \times [0.32, 0.38]$ \\	
    Notes: $m, n$ represent the grid size of the Minesweeper game. $k$ is the total number of mines. $p$ is the probability that a given cell contains a mine (range: $0.32$ to $0.38$).

    \item \textbf{LangFord problem}\\
    Parameter $\Theta$: $\{k\}$ \\
    Dataset Range: $[50, 291] $\\
    Notes: A sequence of $2k$ elements, where numbers from $1$ to $k$ appear twice. The first occurrence of a number $i$ is exactly $i$ positions away from the second occurrence.

    \item \textbf{KnightTour problem} \\
    Parameter $\Theta$: $\{k\}$ \\
    Parameter Space: $[12, 75]$ \\
    Notes: A $k \times k$ chessboard where a knight's tour is attempted, covering all squares and returning to the start point. (A solution is not possible for odd-sized boards, i.e., they are unsatisfiable.)
    \item \textbf{Zamkeller problem}: \\
    Parameter $\Theta$: $\{k, n\}$\\
    Parameter Space: $[3, 34] \times [25, 100]$ \\
    Notes: $k$ represents the total sequence length, and $n$ represents the subsequence length. For all subsequences of length $k$, the goal is to change them into the minimum number of distinct sequences.
\end{itemize}

\section{Prompt Template and Answers}
\subsection{Prompts and Answers in Motivation Questions}
\label{motivation question}

To verify our motivation, we ask LLMs the following questions: 
\begin{itemize}
    \item Can you tell me the definition of SAT?
    \item Can you provide the pseudocode of CDCL solver?
    \item Can you write C++ code based on the pseudocode?
\end{itemize}
Details of answers provided by LLMs are shown below:

\begin{tcolorbox}[colback=blue!10!white, colframe=blue!70!black, coltitle=black, breakable]
Answer of Question 1: Can you tell me the definition of SAT? \\
\begin{lstlisting}[framerule=0pt]
Yes, I understand the Boolean Satisfiability Problem (often abbreviated as SAT). It's a decision problem that asks whether there exists an assignment of truth values to variables that makes a given Boolean formula true. More formally, the SAT problem involves determining if there is a way to assign True or False to variables in a Boolean formula (typically expressed in Conjunctive Normal Form, or CNF) so that the entire formula evaluates to True.

Example:
Consider the Boolean formula:

This formula is satisfiable if there is an assignment of True and False to x_1 and x_2 that makes the entire formula evaluate to True.
One possible satisfying assignment is x_1 = True and x_2 = True.
CNF (Conjunctive Normal Form):
SAT problems are often expressed in CNF, which is a conjunction (AND) of clauses, where each clause is a disjunction (OR) of literals (a variable or its negation). A formula is satisfiable if there is some way to make each clause in the CNF true.

\end{lstlisting}
\end{tcolorbox}

\begin{tcolorbox}[colback=blue!10!white, colframe=blue!70!black, coltitle=black, breakable]
Answer of Question 2: Can you tell me the definition of SAT? \\
\begin{lstlisting}[framerule=0pt]
CDCL_SAT_Solver(CNF_formula):
    initialize assignment_stack = []
    initialize clause_database = CNF_formula
    initialize decision_level = 0

    while True:
        # Perform unit propagation to make implied assignments
        conflict_clause = Unit_Propagation(clause_database, assignment_stack)
        
        if conflict_clause is not None:
            # Conflict occurred, analyze and learn a new clause
            if decision_level == 0:
                # If at decision level 0, unsatisfiable
                return UNSAT
            else:
                # Analyze the conflict and learn a new clause
                learned_clause, backtrack_level = Analyze_Conflict(conflict_clause)
                
                # Add the learned clause to the clause database
                clause_database.add(learned_clause)
                
                # Backjump to the level indicated by conflict analysis
                Backjump(assignment_stack, backtrack_level)
                decision_level = backtrack_level
                
        else:
            # No conflict, check if all variables are assigned
            if All_Variables_Assigned(assignment_stack):
                return SAT  # Satisfiable solution found
            
            # Make a decision: choose a new variable and assign it
            decision_literal = Pick_Unassigned_Variable(assignment_stack)
            decision_level += 1
            assignment_stack.push(decision_literal)

            # Optionally restart the search based on a heuristic
            if Should_Restart():
                Restart_Search(assignment_stack)
                decision_level = 0

# Helper Functions:

Unit_Propagation(clause_database, assignment_stack):
    while there are unit clauses in clause_database:
        assign the unit clause's remaining literal to satisfy the clause
        if a conflict arises:
            return conflict_clause
    return None  # No conflict

Analyze_Conflict(conflict_clause):
    learned_clause = derive a new clause from the conflict (e.g., using resolution)
    backtrack_level = determine the level to backtrack to
    return learned_clause, backtrack_level

Backjump(assignment_stack, backtrack_level):
    remove all assignments in assignment_stack after backtrack_level
    restore the state of the solver at that level

Pick_Unassigned_Variable(assignment_stack):
    choose the next variable to assign (often based on a heuristic like VSIDS or activity-based scoring)
    return decision_literal

Should_Restart():
    return True if the solver should restart (based on a restart policy), otherwise False

Restart_Search(assignment_stack):
    clear the assignment_stack but retain learned clauses in clause_database

\end{lstlisting}
\end{tcolorbox}

\begin{tcolorbox}[colback=blue!10!white, colframe=blue!70!black, coltitle=black, breakable]
Answer of Question 3: Can you write C++ code based on the pseudocode? \\
\begin{lstlisting}[framerule=0pt]
#include <iostream>
#include <vector>
#include <stack>
#include <set>
#include <unordered_map>

using namespace std;

// Represents a clause, which is a disjunction of literals.
typedef vector<Literal> Clause;

class CDCLSolver {
public:
    CDCLSolver(const vector<Clause>& formula) : clauseDatabase(formula), decisionLevel(0) {}

    // Runs the CDCL algorithm on the CNF formula.
    bool solve() {
        while (true) {
            Clause conflictClause = unitPropagation();
            if (!conflictClause.empty()) {
                // Conflict found
                if (decisionLevel == 0) {
                    return false; // UNSAT
                } else {
                    Clause learnedClause;
                    int backtrackLevel;
                    tie(learnedClause, backtrackLevel) = analyzeConflict(conflictClause);
                    clauseDatabase.push_back(learnedClause);
                    backjump(backtrackLevel);
                    decisionLevel = backtrackLevel;
                }
            } else {
                if (allVariablesAssigned()) {
                    return true; // SAT
                }
                // Make a decision
                Literal decisionLiteral = pickUnassignedVariable();
                decisionStack.push(decisionLiteral);
                decisionLevel++;
                assignment[abs(decisionLiteral)] = (decisionLiteral > 0);
            }

            if (shouldRestart()) {
                restartSearch();
            }
        }
    }

private:
    vector<Clause> clauseDatabase; // CNF formula
    stack<Literal> decisionStack;  // Stack of decisions 
    unordered_map<int, bool> assignment;  // Maps variables to their assigned truth values
    int decisionLevel;  // Current decision level in the search

    // Helper function: Perform unit propagation
    Clause unitPropagation() {
        // Simplified unit propagation
        while (true) {
            bool conflict = false;
            for (const Clause& clause : clauseDatabase) {
                int unassignedCount = 0;
                Literal unassignedLiteral = 0;
                bool clauseSatisfied = false;
                for (Literal lit : clause) {
                    int var = abs(lit);
                    if (assignment.find(var) != assignment.end()) {
                        if ((lit > 0 && assignment[var]) || (lit < 0 && !assignment[var])) {
                            clauseSatisfied = true;
                            break;
                        }
                    } else {
                        unassignedCount++;
                        unassignedLiteral = lit;
                    }
                }
                if (!clauseSatisfied && unassignedCount == 0) {
                    return clause; // Conflict found
                } else if (!clauseSatisfied && unassignedCount == 1) {
                    // Unit clause, assign the literal
                    assignment[abs(unassignedLiteral)] = (unassignedLiteral > 0);
                    decisionStack.push(unassignedLiteral);
                }
            }
            if (!conflict) break;
        }
        return {}; // No conflict
    }

    // Helper function: Analyze conflict and learn a new clause
    pair<Clause, int> analyzeConflict(const Clause& conflictClause) {
        Clause learnedClause = conflictClause;  // Simplified conflict analysis (can be improved)
        int backtrackLevel = decisionLevel - 1; // Backtrack one level for simplicity
        return {learnedClause, backtrackLevel};
    }

    // Helper function: Perform backjumping to a specific decision level
    void backjump(int backtrackLevel) {
        while (!decisionStack.empty() && decisionLevel > backtrackLevel) {
            Literal lastDecision = decisionStack.top();
            decisionStack.pop();
            assignment.erase(abs(lastDecision));  // Unassign variables
            decisionLevel--;
        }
    }

    // Helper function: Pick an unassigned variable for decision
    Literal pickUnassignedVariable() {
        for (int i = 1; i <= 10; ++i) {  // Assume variables are from 1 to 10 for simplicity
            if (assignment.find(i) == assignment.end()) {
                return i;  // Return the first unassigned variable
            }
        }
        return 0;  // All variables are assigned 
    }

    // Helper function: Check if all variables are assigned
    bool allVariablesAssigned() {
        for (int i = 1; i <= 10; ++i) {  // Assume variables are from 1 to 10 for simplicity
            if (assignment.find(i) == assignment.end()) {
                return false;
            }
        }
        return true;
    }

    // Helper function: Determine if a restart is necessary
    bool shouldRestart() {
        // Simplified: Always return false (can implement heuristics here)
        return false;
    }

    // Helper function: Restart the search
    void restartSearch() {
        // Reset the assignment and decision stack
        decisionStack = stack<Literal>();
        assignment.clear();
        decisionLevel = 0;
    }
};

int main() {
    vector<Clause> cnfFormula = {{1, 2}, {-1, 3}, {-2, -3}};

    CDCLSolver solver(cnfFormula);
    bool result = solver.solve();

    if (result) cout << "SATISFIABLE" << endl;
    else cout << "UNSATISFIABLE" << endl;

    return 0;
}

\end{lstlisting}
\end{tcolorbox}

\subsection{Prompts and Answers in Experiments }
\label{sec:prompt}
The success of AutoSAT relies heavily on the prompt engineering for LLMs. The prompt templates for AutoSAT are presented here. Brief interpretations are provided below.  
\begin{enumerate}
    \item Define the \textbf{Role} of an agent as a solver expert who needs to assess and improve the heuristics in an SAT solver.
    \item Clearly state the \textbf{Goal}, such as providing optimization suggestions, writing code, or feedback.
    \item
    Enhance the agents' capabilities by inserting optional \textbf{Tips} that guide them to avoid common mistakes during code generation. Additionally, through this flexible interface, agents can effectively utilize external codes and results and can be instructed to specify the types of modification directions such as changing parameters, modifying heuristics, or adding new heuristics.
    \item Total SAT solver code is appended at the end of each prompt to ensure all agents are in the same context.
\end{enumerate}

Maintaining consistency in the prompt design ensures that agents operate in the same context, thereby enabling their outputs to remain coherent. This consistency provides the possibility for the iterative refinement of heuristics in SAT solvers. We also provide actual prompts and answers below.

\begin{tcolorbox}[colback=gray!10!white, colframe=gray!50!black, title=Prompt]
You are a SAT solver researcher trying to rewrite the bump\_var function(s). \\

Your goal is to improve the SAT solver by rewriting the bump\_var function(s), after reading and understanding the <key code> of SAT solver below.\\

Tips:\\
1) Your rewrited function code must start with '''// start {function name}''' and end with '''// end {function name}''' \\
2) Your rewrited function(s) code must be different from original code, not just rewrite code synonymous! \\
3) You are not allowed to create your own new function(s) in the rewrited function(s).  You are not allowed to create your own new global variables, but you can use the global variables existing in the <key code>. \\
4) Make sure the rewrited function(s) code can be executed correctly. \\

<key code> of SAT solver is:\\
...
\end{tcolorbox}

\begin{tcolorbox}[colback=blue!10!white, colframe=blue!70!black, coltitle=black, breakable]
Answer: \\
\begin{lstlisting}[framerule=0pt]

Based on the provided key code of the SAT solver, I will rewrite the `bump_var` function to improve the SAT solver. The goal is to modify the function in a way that is different from the original code, without creating new functions or global variables, and ensuring that the rewritten function can be executed correctly.

Here is the rewritten `bump_var` function:

```cpp
// start bump_var
void Solver::bump_var(int var, double coeff) {
    activity[var] += var_inc * coeff; // Increment activity by the variable increment times the coefficient
    
    // Check if the activity of any variable exceeds a large threshold, and normalize if so.
    const double large_threshold = 1e100;
    bool normalize = false;
    for (int i = 1; i <= vars; i++) {
        if (activity[i] > large_threshold) {
            normalize = true;
            break;
        }
    }
    if (normalize) {
        double max_activity = 0;
        for (int i = 1; i <= vars; i++) {
            if (activity[i] > max_activity) {
                max_activity = activity[i];
            }
        }
        for (int i = 1; i <= vars; i++) {
            activity[i] /= max_activity;
        }
        var_inc /= max_activity;
    }
    
    // Update the variable's position in the heap if it is present
    if (vsids.inHeap(var)) vsids.update(var);
}
// end bump_var
```

In this rewritten version, I've removed the periodic normalization that was previously done every 5000 conflicts. Instead, I've introduced a check that normalizes the activities if any variable's activity exceeds a large threshold (1e100). This approach avoids the need for a periodic check and focuses on the condition that actually necessitates normalization. The normalization process itself is also slightly modified to iterate through the activities twice, first to find the maximum activity and then to normalize, which may be more efficient if normalization is a rare event.
\end{lstlisting}
\end{tcolorbox}

\subsection{EoH Experiments}
\label{sec:eoh-exp}
In this section, we adopt a format similar to works like Funsearch and attempt to generate the entire solver's code using LLMs. The specific prompt template is based on EoH~\shortcite{evolution}, as shown in the EoH prompt. We also provide its answers. The scale of SAT solvers is quite large, but we can find that LLMs only provide code within 1k tokens, which means they are unable to achieve the performance of commonly used SAT solvers. Even for EasySAT, the most lightweight CDCL solver, the code exceeds $5k$ tokens. 
\begin{tcolorbox}[colback=gray!10!white, colframe=gray!50!black, title=EoH Prompt]
I need help designing a SAT solver that efficiently solves Boolean satisfiability problems. The solver should be able to process a CNF formula and return a satisfying assignment if one exists or indicate that the formula is unsatisfiable. The final goal is to optimize the solver's runtime performance and accuracy.\\

Please design a new SAT-solving algorithm.\\
Firstly, describe your new algorithm and its main steps in one sentence.
Next, implement it in Python as a function named solve\_sat. This function should accept one input: cnf\_formula. The function should return one output: result. cnf\_formula is a list of clauses, where each clause is a list of literals represented as integers. The output named result should be a tuple containing a Boolean value indicating satisfiability and a dictionary representing a satisfying assignment if one exists, or an empty dictionary if unsatisfiable.\\

Note that cnf\_formula is a list of lists with integer values indicating literals, and the function should not involve any probabilistic components. Ensure that the algorithm maintains self-consistency and does not provide additional explanations.
\end{tcolorbox}

\begin{tcolorbox}[colback=blue!10!white, colframe=blue!70!black, coltitle=black, breakable]
Answer Code 1: \\
\begin{lstlisting}[framerule=0pt]
#include <vector>
#include <unordered_map>
#include <optional>

bool solve_dpll(const std::vector<std::vector<int>>& cnf_formula, std::unordered_map<int, bool>& assignment) {
    if (cnf_formula.empty()) return true;

    for (const auto& clause : cnf_formula) {
        if (clause.empty()) return false;
    }

    // Choose first unassigned variable (naive implementation)
    int variable = 0;
    for (const auto& clause : cnf_formula) {
        for (int literal : clause) {
            variable = abs(literal);
            if (assignment.find(variable) == assignment.end()) goto found_variable;
        }
    }
found_variable:

    for (int value : {1, -1}) {
        assignment[variable] = (value == 1);

        // Propose assignment with current variable choice
        std::vector<std::vector<int>> new_cnf_formula;
        for (const auto& clause : cnf_formula) {
            bool clause_is_satisfied = false;
            std::vector<int> new_clause;
            for (int literal : clause) {
                if (literal == variable && value == 1) {
                    clause_is_satisfied = true;
                    break;
                } else if (literal == -variable && value == -1) {
                    clause_is_satisfied = true;
                    break;
                } else if (literal != variable && literal != -variable) {
                    new_clause.push_back(literal);
                }
            }
            if (!clause_is_satisfied) new_cnf_formula.push_back(new_clause);
        }

        if (solve_dpll(new_cnf_formula, assignment)) return true;

        assignment.erase(variable);
    }
    return false;
}

std::pair<bool, std::unordered_map<int, bool>> solve_sat(const std::vector<std::vector<int>>& cnf_formula) {
    std::unordered_map<int, bool> assignment;
    bool is_satisfiable = solve_dpll(cnf_formula, assignment);
    return {is_satisfiable, is_satisfiable ? assignment : std::unordered_map<int, bool>{}};
}
\end{lstlisting}
\end{tcolorbox}

\begin{tcolorbox}[colback=blue!10!white, colframe=blue!70!black, coltitle=black, breakable]
Answer Code 2: \\
\begin{lstlisting}[framerule=0pt]
#include <vector>
#include <unordered_map>
#include <unordered_set>
#include <iostream>

bool solve_recursive(std::vector<std::vector<int>>& cnf_formula, std::unordered_map<int, bool>& assignment) {
    // Check if all clauses are satisfied
    bool all_satisfied = true;
    for (const auto& clause : cnf_formula) {
        bool clause_satisfied = false;
        for (int literal : clause) {
            int var = abs(literal);
            bool value = (literal > 0);

            if (assignment.find(var) != assignment.end() && assignment[var] == value) {
                clause_satisfied = true;
                break;
            }
        }
        if (!clause_satisfied) {
            all_satisfied = false;
            break;
        }
    }
    if (all_satisfied) return true;
    // Check if a clause is unsatisfied
    for (const auto& clause : cnf_formula) {
        bool clause_unsatisfied = true;
        for (int literal : clause) {
            int var = abs(literal);
            if (assignment.find(var) == assignment.end()) {
                clause_unsatisfied = false;
                break;
            }
        }
        if (clause_unsatisfied) return false;
    }

    // Choose an unassigned variable
    int variable = 0;
    for (const auto& clause : cnf_formula) {
        for (int literal : clause) {
            int var = abs(literal);
            if (assignment.find(var) == assignment.end()) {
                variable = var;
                break;
            }
        }
        if (variable != 0) break;
    }

    assignment[variable] = true;
    if (solve_recursive(cnf_formula, assignment)) return true;
    assignment[variable] = false;
    if (solve_recursive(cnf_formula, assignment)) return true;
    assignment.erase(variable);
    return false;
}

std::pair<bool, std::unordered_map<int, bool>> solve_sat(std::vector<std::vector<int>>& cnf_formula) {
    std::unordered_map<int, bool> assignment;
    bool is_satisfiable = solve_recursive(cnf_formula, assignment);
    return { is_satisfiable, is_satisfiable ? assignment : std::unordered_map<int, bool>() };
}

int main() {
    std::vector<std::vector<int>> cnf_formula = { {1, -3, 4}, {-1, 2, 3}, {-2, -3}, {2, -4} };
    auto result = solve_sat(cnf_formula);
    std::cout << "Satisfiable: " << result.first << std::endl;
    if (result.first) {
        std::cout << "Assignment: ";
        for (const auto& assign : result.second) {
            std::cout << assign.first << "=" << assign.second << " ";
        }
        std::cout << std::endl;
    }
    return 0;
}
\end{lstlisting}
\end{tcolorbox}

\begin{tcolorbox}[colback=blue!10!white, colframe=blue!70!black, coltitle=black, breakable]
Answer Code 3: \\
\begin{lstlisting}[framerule=0pt]
#include <iostream>
#include <vector>
#include <unordered_map>
#include <optional>
using namespace std;

struct SATResult {
    bool satisfiable;
    unordered_map<int, bool> assignment;
};

optional<unordered_map<int, bool>> DPLL(const vector<vector<int>>& cnf_formula, unordered_map<int, bool>& assignment) {
    for (const auto& clause : cnf_formula) {
        bool clause_satisfied = false;
        bool has_unassigned_literal = false;
        for (int literal : clause) {
            auto it = assignment.find(abs(literal));
            if (it != assignment.end() && it->second == (literal > 0)) {
                clause_satisfied = true;
                break;
            }
            if (it == assignment.end()) {
                has_unassigned_literal = true;
            }
        }
        if (!clause_satisfied && !has_unassigned_literal) {
            return {}; // Clause unsatisfiable and no unassigned literal
        }
    }
    
    int unassigned_var = 0;
    for (const auto& clause : cnf_formula) {
        for (int literal : clause) {
            if (assignment.find(abs(literal)) == assignment.end()) {
                unassigned_var = abs(literal);
                break;
            }
        }
        if (unassigned_var != 0) break;
    }
    
    if (unassigned_var == 0)  return assignment;
    
    assignment[unassigned_var] = true;
    auto result = DPLL(cnf_formula, assignment);
    if (result.has_value()) {
        return result;
    }
    assignment[unassigned_var] = false;
    return DPLL(cnf_formula, assignment);
}

SATResult solve_sat(const vector<vector<int>>& cnf_formula) {
    unordered_map<int, bool> assignment;
    auto result = DPLL(cnf_formula, assignment);
    if (result.has_value()) {
        return {true, result.value()};
    }
    return {false, {}};
}

int main() {
    vector<vector<int>> cnf_formula = {{1, -3}, {-1, 2, 3}, {-1, -2}};
    SATResult result = solve_sat(cnf_formula);
    
    cout << "Satisfiable: " << result.satisfiable << endl;
    if (result.satisfiable) {
        cout << "Assignment: ";
        for (const auto& [var, val] : result.assignment) {
            cout << var << ": " << val << ", ";
        }
        cout << endl;
    }
    return 0;
}
\end{lstlisting}
\end{tcolorbox}

\end{document}